\documentclass[conference]{IEEEtran}
\usepackage{amsmath,amsfonts}
\usepackage{tabularx}
\usepackage{algorithm}
\usepackage{algpseudocode}
\usepackage{orcidlink}
\usepackage{physics}
\usepackage[shortlabels]{enumitem}
\usepackage{array}
\hypersetup{%
    pdfborder = {0 0 0}
}
\usepackage{textcomp}
\usepackage{stfloats}
\usepackage{url}
\usepackage{verbatim}
\usepackage{graphicx}
\usepackage{cite}
\begin{document}

\title{Quantum-Aided Active Device Detection in Energy-Harvesting Symbiotic Radio Networks}

\author{
\IEEEauthorblockN{
Remon Polus\orcidlink{0000-0002-5527-0265},
Deemah H. Tashman\orcidlink{0000-0002-3054-5445},
and
Soumaya Cherkaoui\orcidlink{0000-0001-6140-770X}
}
\IEEEauthorrefmark{1}
Department of Computer and Software Engineering,
Polytechnique Montréal, Montréal, Canada \\
Email: \{remon.polus, deemah.tashman, soumaya.cherkaoui\}@polymtl.ca
}

\maketitle

\begin{abstract}
Massive connectivity in next-generation networks demands energy- and spectrum-efficient solutions for large-scale Internet of Things (IoT) deployments.
Symbiotic radio (SR) enables passive IoT devices to communicate by backscattering existing cellular transmissions.
A key challenge in uplink SR is active device detection (ADD), which directly affects decoding reliability, interference management, and system throughput.
We propose an energy-harvesting code-domain non-orthogonal multiple access (NOMA)-SR system in which IoT devices harvest energy from ambient uplink signals and backscatter information using low-density spreading (LDS) codes.
To reduce the complexity of ADD, Grover's quantum search algorithm is employed, providing a quadratic reduction in oracle-query complexity over exhaustive maximum-likelihood (ML) search.
Numerical results show that the proposed approach closely approaches ML performance while substantially reducing the number of search iterations, demonstrating its potential for scalable ambient IoT systems.
\end{abstract}


\section{Introduction}
\label{S1}
With Industry 4.0 evolving toward the fifth industrial revolution, the number of Internet of Things (IoT) devices is projected to reach 29 billion worldwide by 2030~\cite{de2021survey}. 
The International Telecommunication Union (ITU) IMT-2030 framework envisions sixth-generation (6G) wireless networks capable of supporting ultra-massive and seamless connectivity.
However, the rapid proliferation of IoT devices is expected to exacerbate energy consumption and spectrum scarcity~\cite{wang2025energy}, motivating the development of energy- and spectrum-efficient communication technologies.
Symbiotic radio (SR) has emerged as a promising paradigm for addressing these challenges~\cite{long2019full}.
By enabling passive IoT devices, referred to as backscatter devices, to modulate and reflect ambient radio-frequency signals, SR reduces the reliance on active radio-frequency (RF) components while providing additional multipath diversity to the primary communication system~\cite{mondal2025comprehensive}.
This mutually beneficial coexistence between passive IoT devices and cellular infrastructure makes SR a promising technology for supporting ambient IoT in 6G networks~\cite{liang2022symbiotic}.

The integration of energy harvesting (EH) and non-orthogonal multiple access (NOMA) provides a viable framework for enabling large-scale energy-constrained SR networks~\cite{moloudian2024rf}.
EH enables IoT devices to extract energy from ambient sources and store it for subsequent communication and computation, thereby reducing dependence on conventional batteries and associated replacement costs~\cite{kim2025challenges}.
In parallel, NOMA improves spectral efficiency by allowing multiple devices to concurrently access the same time-frequency resources~\cite{dai2018survey}.
In code-domain NOMA (CD-NOMA), users are distinguished through device-specific non-orthogonal spreading sequences characterized by low cross-correlation, with low-density spreading (LDS) and sparse code multiple access (SCMA) constituting representative schemes~\cite{jafarkhani2024modulation}.

A fundamental challenge in uplink SR networks is active device detection (ADD), where the base station (BS) must identify, in real time, the subset of devices that are actively transmitting \cite{habibie2024quantum}.
Accurate ADD is essential for subsequent multi-user decoding, interference management, and throughput optimization.
To address this issue, Grover's quantum search algorithm provides a promising approach to reducing the search complexity of ADD by offering a quadratic speedup for unstructured search problems \cite{botsinis2018quantum}.
However, to the best of the authors' knowledge, quantum-assisted ADD has not yet been investigated in SR networks, where ADD is further affected by energy-harvesting constraints, backscatter channel conditions, and multi-user interference.
This paper addresses this gap by developing a quantum-assisted ADD framework for energy-harvesting SR networks.
The main contributions are as follows:
\begin{itemize}
\item We propose an energy-harvesting CD-NOMA-SR system where IoT devices harvest energy from cellular uplink signals and backscatter information to the BS via LDS codes, enabling battery-less massive IoT connectivity.
\item We apply Grover's quantum search algorithm to ADD in the proposed uplink SR network under Rayleigh fading and AWGN, demonstrating its potential as a low-complexity detection mechanism for large-scale SR networks.
\end{itemize}
The remainder of this paper is organized as follows: Section~\ref{S2} presents the system model for the energy harvesting-based SR network, Section~\ref{S3} details the application of Grover's algorithm to the ADD process, Section~\ref{S4} presents simulation results and discussion, and Section~\ref{S5} concludes the paper and outlines future work.
\section{System Model}
\label{S2}
\begin{figure}[t]
\vspace{3pt}
\centering
\includegraphics[width=0.7\columnwidth]{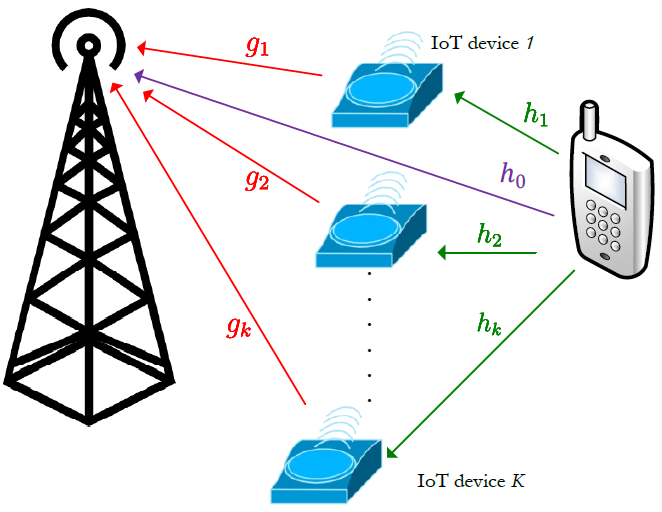}
\caption{System model of an uplink SR energy-harvesting network.}
\label{fig_01}
\end{figure}

As illustrated in Fig.~\ref{fig_01}, we consider an SR system in which a cellular user communicates with the BS while $K$ IoT devices harvest energy from the uplink transmission to support battery-less operation.
Once sufficiently charged, the IoT devices become active and backscatter their information using distinct LDS codes, which the BS jointly decodes alongside the cellular transmission.
The cellular user also benefits, as the backscattered signals create additional propagation paths that improve diversity and reliability at the BS.
The frame duration $T$ is divided into an energy-harvesting phase $\tau_0$ and a data-transmission phase $T-\tau_0$.

\subsection{Energy Harvesting}
During $\tau_0$, the signal received at IoT device $k$ is \cite{liang2020symbiotic}
\begin{equation}
\label{eq_1}
x_k(mN+n) = \sqrt{p}\, h_k s(mN+n),
\end{equation}
where $p$ is the cellular transmit power, $h_k$ the channel to device $k$, $s(mN+n)$ the $n^\text{th}$ chip of the $m^\text{th}$ symbol, and $N$ the LDS code length. The power harvested by device $k$ is \cite{tashman2025quantum}
\begin{equation}
\label{eq_2}
p_k = \eta p |h_k|^2,
\end{equation}
with $\eta$ the harvesting efficiency, and the energy accumulated over $\tau_0$ is
\begin{equation}
\label{eq_3}
E_k = p_k \tau_0 = \eta \tau_0 p |h_k|^2.
\end{equation}
During $T-\tau_0$, each device consumes
\begin{equation}
\label{eq_4}
E_c = \frac{T-\tau_0}{T_b} N E_0 + (T-\tau_0)p_c,
\end{equation}
where the first term is the energy for reflection-coefficient adjustment ($E_0$) and the second is the circuit consumption ($p_c$), with $T_b$ the backscatter bit duration \cite{liang2020symbiotic}. Device $k$ activates only if $E_k \geq E_c$, giving the binary activation vector $\mathbf{a} = [a_1,\ldots,a_K]^{\mathsf{T}}$ with
\begin{equation}
\label{eq_6}
a_k =
\begin{cases}
1, & \eta \tau_0 p |h_k|^2 \geq E_c, \\[0.5ex]
0, & \text{otherwise}.
\end{cases}
\end{equation}
After harvesting sufficient energy, active devices begin transmitting by backscattering the cellular signal; to avoid mutual interference, each device multiplies its symbol by a unique LDS code.

\subsection{LDS Codes}
Let $\mathbf{F}\in\{0,1\}^{N\times K}$ denote the LDS factor matrix, where $N$ is the number of chips per codeword, $K$ is the number of available codewords, and each codeword has exactly $J$ nonzero chips; $f_{n,k}=1$ indicates that chip $n$ is nonzero in codeword $k$, and $f_{n,k}=0$ otherwise. The total number of available codewords is given by the binomial coefficient
\begin{equation}
U=\binom{N}{J}.
\end{equation}
For example, with $N=4$, $J=2$,
\begin{equation}
U=\binom{4}{2} = \frac{4!}{2!\,(4-2)!}=6,
\end{equation}
so the system accommodates up to $6$ devices, each assigned a unique codeword, giving the factor matrix
\begin{equation}
\mathbf{F}=
\begin{bmatrix}
1 & 1 & 1 & 0 & 0 & 0 \\
1 & 0 & 0 & 1 & 1 & 0 \\
0 & 1 & 0 & 1 & 0 & 1 \\
0 & 0 & 1 & 0 & 1 & 1
\end{bmatrix},
\end{equation}
where each column has exactly $J=2$ nonzero entries and each row has $d_f=3$ nonzero entries, meaning every chip is shared among $3$ devices. The resulting overloading factor is
\begin{equation}
\lambda = \frac{U}{N} = \frac{6}{4}=150\%.
\end{equation}
\subsection{Received Signal at the BS}
The received signal at the BS comprises both the direct-link signal from the cellular user and the backscatter links from all active IoT devices, which can be expressed as
\begin{equation}
\label{eq_7}
\begin{split}
y(mN+n) &= \sqrt{p}\, h_0 s(mN+n) \\ 
&\quad + \sum_{k=1}^{K} \sqrt{p}\, h_k g_k b_k(m) a_k c_k(n) s(mN+n) \\ 
&\quad + u(mN+n),
\end{split}
\end{equation}
where $h_0$ is the channel coefficient of the direct link from the cellular device to the BS.
$g_k$ denotes the channel coefficient of the channel between IoT device $k$ to the BS.
We assume that all these channels follow Rayleigh distribution.
$b_k(m)$ is the symbol transmitted by device $k$ and $c_k(n)$ is the LDS code used by device $k$.
$u(mN+n) \sim \mathcal{CN}(0,\sigma^2)$ is the AWGN at the BS, where the total signal-to-noise ratio (SNR) can be defined as
\begin{equation}
\label{eq_8}
\gamma=\frac{p|h_0|^2+\sum_{k=1}^{K}a_k p |g_k h_{k}|^2}{\sigma^2}.
\end{equation}
Since the direct-link signal is typically much stronger than the backscattered signals, the receiver first decodes $s(mN+n)$ and removes it using successive interference cancellation (SIC) \cite{wang2025energy}.
After SIC, the resulting signal is given as
\begin{equation}
\label{eq_9}
\begin{split}
\tilde{y}(mN+n) &= \sum_{k=1}^{K} \sqrt{p}\, g_k h_k a_k c_k(n) s(mN+n) b_k(m) \\ 
&\quad + u(mN+n).
\end{split}
\end{equation}
\section{Grover's Algorithm for Active IoT Device Detection}
\label{S3}
At the BS, the goal is to detect the active IoT devices, corresponding to the ADD problem, whose search space grows as $2^K$, rendering classical exhaustive detection prohibitive for large-scale SR deployments.
We instead employ a quantum search framework that exploits superposition to evaluate multiple candidate activity patterns in parallel, reducing the computational burden of ADD \cite{8972916}.
Among quantum search methods, Grover's algorithm is widely used for finding a target element in an unsorted database of size $B$, offering a quadratic speedup of $\mathcal{O}(\sqrt{B})$ over the classical $\mathcal{O}(B)$ \cite{botsinis2018quantum}. It is built on two main components: the Oracle and the Diffuser.

The Oracle identifies the quantum state associated with the desired outcome — in the ADD context, the codeword that best matches the received superposed signal from the active backscatter IoT devices — by marking it via a phase inversion.
Using an auxiliary qubit initialized in $|-\rangle$, the Oracle operator $O_{\nu}$ applies a phase shift of $-1$ to a state $|x\rangle$ satisfying $Y(x)=\nu$, leaving all other states unchanged \cite{botsinis2018quantum}:
\begin{IEEEeqnarray}{lCr} 
O_\nu|x\rangle= \begin{cases}
   -|x\rangle   , & \text{if} \; Y(x)=\nu  \\
   |x\rangle  , &  \text{if} \; Y(x) \neq \nu 
 \end{cases}.
\end{IEEEeqnarray}

While the Oracle tags the desired state, the Diffuser amplifies it via inversion-about-the-mean, and iterating the two increases the probability of observing the correct solution \cite{botsinis2018quantum}. The number of Grover iterations follows \cite{tashman2025quantum}
\begin{IEEEeqnarray}{lcr}
\kappa=\lfloor \frac{\pi}{4} \sqrt{\frac{B}{L}} \rfloor,
\end{IEEEeqnarray}
where $B=2^K$ is the search space size (all activity combinations) and $L$ is the number of feasible solutions.

Prior to encoding, the received signal is discretized to ensure compatibility with binary quantum circuits. Assuming unipolar signaling with symbols in $\{0,1\}$, the received signal $y$ is quantized as \cite{tashman2025quantum}
\begin{IEEEeqnarray}{lcr}
y_I=\min\left(\max\left(0,\mathrm{round}(y)\right),2^s-1\right),
\end{IEEEeqnarray}
where $s$ sets the number of quantization levels.

\subsection{Average Probability of Success}
The average probability of success $P_s$ measures the BS's ability to correctly distinguish active from inactive devices, with higher $P_s$ implying a lower false-positive rate. Letting $\hat{a}_k$ denote the estimated activity of device $k$, detection is successful if
\begin{equation}
r_k = 
\begin{cases}
1, & \hat{a}_k = a_k, \\
0, & \text{otherwise,}
\end{cases} 
\quad k = 1,\ldots,K,
\end{equation}
and the average success probability over $Z_\text{total}$ trials is
\begin{equation}
P_s = \frac{\sum_{t=1}^{Z_\text{total}} \sum_{k=1}^{K} r_k^{(t)}}{K \, Z_\text{total}}.
\end{equation}


\section{Numerical Results}
\label{S4}
This section evaluates the performance of the proposed approach.
Since the quantum algorithm is implemented using a classical simulator, the evaluation can be computationally intensive \cite{piron2024quantum}.
Therefore, a simplified network architecture is considered to enable practical evaluation.
The simulation parameters are summarized in Table~\ref{Tab I}.
\begin{table}[ht!]
\centering
\caption{Simulation Parameters}
\fontsize{10}{12}\selectfont 
\begin{tabular}{|>{\centering\arraybackslash}p{4cm}||>{\centering\arraybackslash}p{1.5cm}|>{\centering\arraybackslash}p{1.75cm}|}
\hline
\textbf{Parameter} & \textbf{Symbol} & \textbf{Value} \\
\hline
Bit duration & $T_b$ & $0.02$ ms \\
\hline
Total time slot & $T$ & 0.1 s \\
\hline
Energy harvesting slot & $\tau_0$ & 0.01 s \\
\hline
Energy required to adjust the reflection coefficient & $E_0$ & 10 nJ \\
\hline
Harvesting efficiency & $\eta$ & 0.9 \\
\hline
Noise variance & $\sigma^2$ & 0.005 \\
\hline
Circuit power & $p_c$ & -5 dBm \\
\hline
Number of Iterations & - & $10^4$ \\
\hline
\end{tabular}
\label{Tab I}
\end{table}

\begin{figure}[t]
  \vspace{3pt}
\centering
\includegraphics[width=0.80\columnwidth]{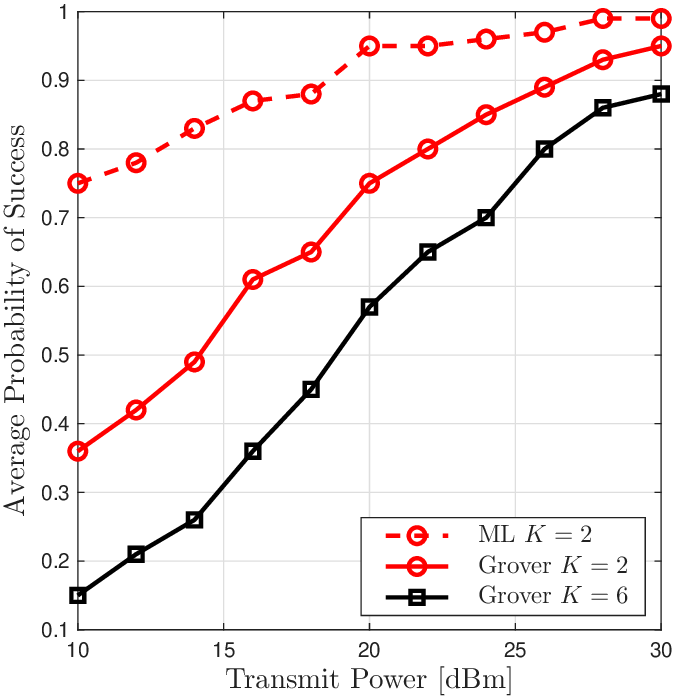}
\caption{Average probability of success versus the transmit power in dBm (varying $K$).}
\label{Res_01}
\end{figure}

Fig.~\ref{Res_01} shows the average probability of successful detection versus transmit power from 10 to 30 dBm for classical maximum-likelihood (ML) detection, used as the benchmark, and Grover-based detection with $K=2$ and $K=6$, with $J=2$.
As transmit power increases, all schemes exhibit improved detection performance due to the increased received SNR at the BS.
For $K=2$, the Grover-based detector closely approaches the ML benchmark while requiring quadratically fewer search iterations.
As $K$ increases to 6, the search space expands from $2^2=4$ to $2^6=64$ states, while increased multi-user interference leads to some performance degradation.
Nevertheless, the $K=6$ case maintains the expected improvement with transmit power, demonstrating the applicability of Grover-based detection to larger device sets.

\begin{table}[t]
\centering
\caption{Computational Complexity: Grover's Algorithm vs.\ ML Detection}
\label{tab:complexity}
\fontsize{10}{12}\selectfont 
\begin{tabular}{ccccc}
\hline
$K$ & Search space & Grover & ML & Speedup \\
\hline
2 & 4 & 1 & 4 & $4.0\times$ \\
4 & 16 & 3 & 16 & $5.3\times$ \\
6 & 64 & 12 & 64 & $5.3\times$ \\
\hline
\end{tabular}
\end{table}

The corresponding search complexity of the detection schemes in Fig.~\ref{Res_01} is presented in Table~\ref{tab:complexity} that compares the number of oracle calls required by the proposed Grover-based detector with the exhaustive comparisons required by the classical ML benchmark for $K=2,4,6$ IoT devices.
As $K$ increases, the ML search space grows exponentially as $B=2^K$, whereas the number of Grover iterations $\kappa=\left\lfloor\frac{\pi}{4}\sqrt{B}\right\rfloor$ grows as $\mathcal{O}(\sqrt{B})$.
For $K=6$, Grover requires only 12 oracle calls compared with 64 ML comparisons, corresponding to a $5.3\times$ reduction in search operations.
This advantage becomes increasingly significant as the number of IoT devices grows, highlighting the potential of the proposed approach for large-scale SR deployments.

\begin{figure}[t]
  \vspace{3pt}
\centering
\includegraphics[width=0.80\columnwidth]{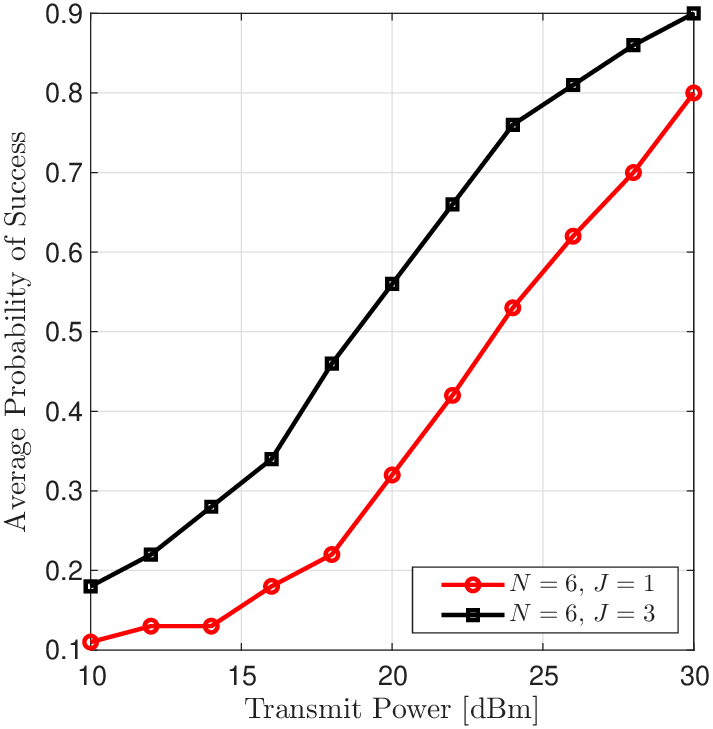}
\caption{Average probability of success versus the transmit power in dBm (varying $J$).}
\label{Res_02}
\end{figure}

Fig.~\ref{Res_02} shows the average probability of successful detection versus transmit power from 10 to 30 dBm for $N=6$, $J=1$ and $J=3$, with $K=6$ IoT devices.
The $J=3$ configuration consistently outperforms $J=1$ due to the greater spreading diversity obtained by allocating multiple chips to each device.
This increased diversity improves the distinguishability of the received backscatter signals and, consequently, detection reliability.
In contrast, $J=1$ provides limited diversity, resulting in weaker signal distinguishability despite lower inter-device interference.
These results demonstrate that increasing the number of chips per device can improve the performance of Grover-based ADD.
\section{Conclusion}
\label{S5}
This paper proposed an energy-harvesting CD-NOMA symbiotic radio framework for battery-less IoT connectivity, where IoT devices harvest energy from ambient cellular uplink signals and backscatter information using LDS codes.
Grover's quantum search algorithm is employed at the BS to identify active devices with reduced search complexity.
Simulation results show that the proposed detector closely approaches the ML upper bound while substantially reducing the number of search iterations.
Future work will investigate implementation on near-term quantum hardware and extensions to multi-user primary networks and larger-scale IoT deployments.


\vfill


\begin{thebibliography}{10}
\providecommand{\url}[1]{#1}
\csname url@samestyle\endcsname
\providecommand{\newblock}{\relax}
\providecommand{\bibinfo}[2]{#2}
\providecommand{\BIBentrySTDinterwordspacing}{\spaceskip=0pt\relax}
\providecommand{\BIBentryALTinterwordstretchfactor}{4}
\providecommand{\BIBentryALTinterwordspacing}{\spaceskip=\fontdimen2\font plus
\BIBentryALTinterwordstretchfactor\fontdimen3\font minus \fontdimen4\font\relax}
\providecommand{\BIBforeignlanguage}[2]{{%
\expandafter\ifx\csname l@#1\endcsname\relax
\typeout{** WARNING: IEEEtran.bst: No hyphenation pattern has been}%
\typeout{** loaded for the language `#1'. Using the pattern for}%
\typeout{** the default language instead.}%
\else
\language=\csname l@#1\endcsname
\fi
#2}}
\providecommand{\BIBdecl}{\relax}
\BIBdecl

\bibitem{de2021survey}
C.~De~Alwis, A.~Kalla, Q.-V. Pham, P.~Kumar, K.~Dev, W.-J. Hwang, and M.~Liyanage, ``{Survey on 6G Frontiers: Trends, Applications, Requirements, Technologies and Future Research},'' \emph{IEEE Open J. Commun. Soc.}, vol.~2, pp. 836--886, 2021.

\bibitem{wang2025energy}
J.~Wang, Z.~Zhao, and Y.-C. Liang, ``{Energy Harvesting-Data Transmission Tradeoff in Symbiotic Radios for Ambient IoT},'' \emph{IEEE Trans. Wireless Commun.}, vol.~25, pp. 1437--1450, 2026.

\bibitem{long2019full}
R.~Long, H.~Guo, L.~Zhang, and Y.-C. Liang, ``{Full-Duplex Backscatter Communications in Symbiotic Radio Systems},'' \emph{IEEE Access}, vol.~7, pp. 21\,597--21\,608, 2019.

\bibitem{mondal2025comprehensive}
S.~Mondal, D.~Bepari, A.~Chandra, K.~Singh, C.-P. Li, and Z.~Ding, ``{A Comprehensive Survey on NOMA-Based Backscatter Communication for IoT Applications},'' \emph{IEEE Internet Things J.}, vol.~12, no.~12, pp. 18\,929--18\,953, 2025.

\bibitem{liang2022symbiotic}
Y.-C. Liang, R.~Long, Q.~Zhang, and D.~Niyato, ``{Symbiotic Communications: Where Marconi Meets Darwin},'' \emph{IEEE Wirel. Commun.}, vol.~29, no.~1, pp. 144--150, 2022.

\bibitem{moloudian2024rf}
G.~Moloudian, M.~Hosseinifard, S.~Kumar, R.~B. Simorangkir, J.~L. Buckley, C.~Song, G.~Fantoni, and B.~O’Flynn, ``{RF Energy Harvesting Techniques for Battery-Less Wireless Sensing, Industry 4.0, and Internet of Things: A Review},'' \emph{IEEE Sens. J.}, vol.~24, no.~5, pp. 5732--5745, 2024.

\bibitem{kim2025challenges}
Y.~Kim, L.~Liu, K.~Takeda, B.~Han, V.~Vintola, C.~Wei, P.~Gupta, C.~Zhang, Z.~Fan, K.~Mukkavilli \emph{et~al.}, ``{Challenges and Advances in Ambient IoT within 3GPP},'' \emph{IEEE Commun. Mag.}, vol.~64, no.~1, pp. 182--188, 2026.

\bibitem{dai2018survey}
L.~Dai, B.~Wang, Z.~Ding, Z.~Wang, S.~Chen, and L.~Hanzo, ``{A Survey of Non-Orthogonal Multiple Access for 5G},'' \emph{IEEE Commun. Surv. Tutor.}, vol.~20, no.~3, pp. 2294--2323, 2018.

\bibitem{jafarkhani2024modulation}
H.~Jafarkhani, H.~Maleki, and M.~Vaezi, ``Modulation and coding for NOMA and RSMA,'' \emph{Proc. IEEE}, vol.~112, no.~9, pp.~1179--1213, 2024.

\bibitem{habibie2024quantum}
M.~I. Habibie, C.~Goursaud, and J.~Hamie, ``{Quantum Minimum Searching Algorithms for Active User Detection in Wireless IoT Networks},'' \emph{IEEE Internet Things J.}, vol.~11, no.~12, pp. 22\,603--22\,615, 2024.

\bibitem{botsinis2018quantum}
P.~Botsinis, D.~Alanis, Z.~Babar, H.~V. Nguyen, D.~Chandra, S.~X. Ng, and L.~Hanzo, ``{Quantum Search Algorithms for Wireless Communications},'' \emph{IEEE Commun. Surv. Tutor.}, vol.~21, no.~2, pp. 1209--1242, 2019.

\bibitem{liang2020symbiotic}
Y.-C. Liang, Q.~Zhang, E.~G. Larsson, and G.~Y. Li, ``{Symbiotic Radio: Cognitive Backscattering Communications for Future Wireless Networks},'' \emph{IEEE Trans. Cogn. Commun. Netw.}, vol.~6, no.~4, pp. 1242--1255, 2020.

\bibitem{tashman2025quantum}
D.~H. Tashman and S.~Cherkaoui, ``{Quantum-Aided Active User Detection for Energy-Efficient CD-NOMA in Cognitive Radio Networks},'' in \emph{Proc. Int. Wirel. Commun. Mob. Comput. Conf. (IWCMC)}.\hskip 1em plus 0.5em minus 0.4em\relax IEEE, 2025, pp. 1661--1666.

\bibitem{8972916}
Y.~Li, M.~Tian, G.~Liu, C.~Peng, and L.~Jiao, ``{Quantum Optimization and Quantum Learning: A Survey},'' \emph{IEEE Access}, vol.~8, pp. 23\,568--23\,593, 2020.

\bibitem{piron2024quantum}
R.~Piron and C.~Goursaud, ``{Quantum Annealing for Active User Detection in NOMA Systems},'' in \emph{Proc. Asilomar Conf. Signals Syst. Comput.}\hskip 1em plus 0.5em minus 0.4em\relax IEEE, 2024, pp. 1448--1452.

\end{thebibliography}
\end{document}